\documentclass{article}

\PassOptionsToPackage{numbers, compress}{natbib}

\usepackage[preprint]{neurips_2019}

\usepackage[utf8]{inputenc} 
\usepackage[T1]{fontenc}    
\usepackage{url}            
\usepackage{booktabs}       
\usepackage{amsfonts}       
\usepackage{nicefrac}       
\usepackage{microtype}      
\usepackage{graphicx}
\usepackage{subcaption}
\usepackage{amsmath}
\usepackage{enumitem}

\makeatletter
\newcommand*\bigcdot{\mathpalette\bigcdot@{.5}}
\newcommand*\bigcdot@[2]{\mathbin{\vcenter{\hbox{\scalebox{#2}{$\m@th#1\bullet$}}}}}
\makeatother

\title{Data-free Knowledge Distillation for Segmentation using Data-Enriching GAN}

\author{
  Kaushal Santosh Bhogale \\
  Department of Computational and Data Sciences\\
  Indian Institute of Science, Bangalore \\
  \texttt{kaushalb@iisc.ac.in} \\
}

\begin{document}

\maketitle

\begin{abstract}
Distilling knowledge from huge pretrained networks to improve performance of tiny networks has favoured deep learning models to be used in many real-time and mobile applications. Several approaches which demonstrate success  in this field have made use of the true training dataset to extract relevant knowledge. In absence of the true dataset however, extracting knowledge from deep networks is still a challenge. Recent works on data-free knowledge distillation demonstrate such techniques on classification tasks. To this end, we explore the task of data-free knowledge distillation for segmentation tasks. First, we identify several challenges specific to segmentation. We make use of the DeGAN training framework to propose a novel loss function for enforcing diversity in a setting where a few classes are underrepresented. Further, we explore a new training framework for performing knowledge distillation in a data-free setting. We get an improvement of 6.93\% in Mean IoU over previous approaches.

\end{abstract}

\section{Introduction}

\begin{figure}
  \centering
  \includegraphics[scale=0.4]{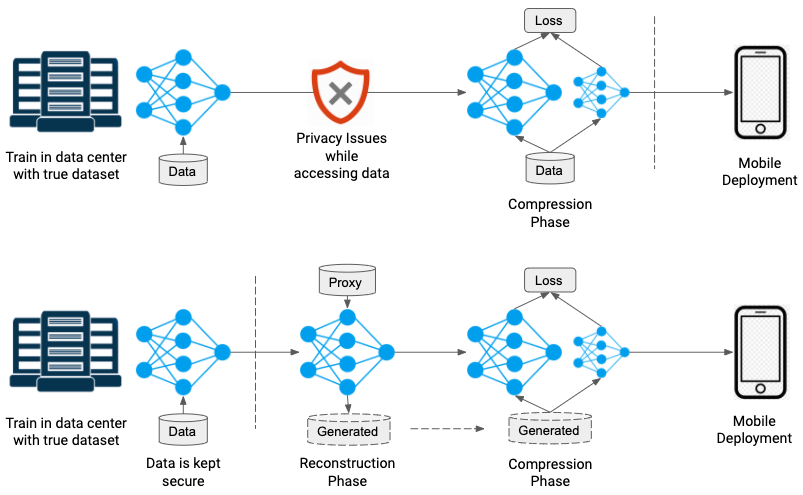}
  \caption{
  \textit{Deployment Pipelines} (Top row) A deployment pipeline where a large model is initially trained on some dataset. Later for mobile deployment when the model needs to be compressed, the dataset needs to be accessed again causing privacy concerns.
  (Bottom row) We present a deployment pipeline where no access to data is required. Instead we use a proxy dataset to generate representative samples to perform model compression.
  }
  \label{fig:banner}
\end{figure}

The two elements that promise the success of deep learning are lots of data and huge networks. An enticing question that still remains unanswered is - \emph{Can we train tiny deep neural networks with a small set of unlabelled images while still maintaining high performance?} Convolutional Neural Networks have shown great success \cite{Krizhevsky2017, He2016, Szegedy2015} in scenarios where large amounts of labelled data is available, i.e. a supervised approach. A part of this success is attributed to the availability of large-scale datasets like ImageNet \cite{Deng2009}. But for dense prediction tasks like segmentation, labelling of datasets is expensive. This restricts the task of creating large densely labelled datasets to big corporations or businesses. Although due to privacy concerns or legal issues these datasets cannot be made public (Figure \ref{fig:banner}), the open-source culture has encouraged many organisations to publish their trained models for public use.

Even if trained models are available, downstream tasks like Knowledge Distillation \cite{Kdoriginal} require the need for labelled datasets for training tiny models. Recent works on Data-free Knowledge Distillation \cite{nayak2019zero, lopes2017data}  make an attempt to achieve compression without access to any large datasets. These approaches have shown to work well on classification tasks. Owing to the success of data-free knowledge distillation for classification tasks, a question arises - \emph{Are these techniques extendable to other computer vision tasks like segmentation?} 

Convolutional Neural Networks are able to achieve high performance on classification tasks due to local invariances learnt by down-sampling and pooling layers of the network. On the other hand, networks for segmentation aim to improve localization accuracy. To achieve scale invariance, these networks also view objects at multiple levels \cite{he2015spatial}. To improve object boundaries, fully connected CRF \cite{chen2017deeplab} is also employed. Many state-of-the-art architectures for segmentation are fully convolutional. These idiosyncrasies make segmentation networks different from classification networks. Thus, direct application of ideas which work on classification networks fail to perform well on segmentation tasks.

The primary challenge in data-free knowledge distillation is to be able to generate samples which closely match the distribution of the data on which the teacher network was trained. To do so, we generate samples which look like real images, which confidently get classified by the network and which are diverse enough to represent all classes of the task. We empirically show that these conditions are sufficient for generating representative samples for downstream tasks. To our best knowledge, this is the first work which tries to tackle the task of data-free knowledge distillation for segmentation.

This work has previously been explored for classification tasks by \citet{addepalli2019degan}. The main contribution of this project is to extend the ideas to segmentation tasks. We identify the following three challenges and propose solutions to them.

\begin{enumerate}[noitemsep]
    \item Generation of images of large sizes.
    \item Handling the unbalanced nature of class distributions in segmentation tasks.
    \item Ensuring both diversity and quality of samples used for knowledge distillation.
\end{enumerate}

\section{Background}

\subsection{Semantic segmentation}
The task of Semantic Segmentation has widely used applicability in self-driving cars \cite{treml2016speeding}, autonomous drones, etc. With advent of large datasets for segmentation like Pascal VOC \cite{everingham2010pascal}, urban scenes datasets like Cityscapes \cite{cordts2016cityscapes}, CamVid \cite{brostow2009semantic}; development of end-to-end deep learning models has been made possible. Fully convolutional Networks\cite{long2015fully} have paved a path which all modern segmentation networks use. From there on, many novel architectures and loss formulations have been proposed to make these systems better. \citet{chen2017rethinking} proposed the Deeplabv3 architecture which uses atrous convolutions and pyramid pooling to improve performance. Their architecture consists of an ASPP module which can be used with any backbone network architectures like ResNet or Mobilenet as a feature extractor. We use DeepLabV3 architecture for segmentation in our experiments.

\subsection{Knowledge Distillation}
To improve accuracy of deep neural networks, the models keep getting larger in size. Larger models (called Teacher Network) are known to learn better representations than smaller models (called Student Network) when trained on a given dataset. The teacher network is overparameterized, thus the knowledge distillation framework allows student network to leverage from the dark knowledge learnt by a teacher network to improve its performance. \citet{Kdoriginal} proposed to distill this knowledge by minimizing the distribution with softer probabilities. Several works make use of intermediate representations \cite{romero2014fitnets} instead of softmax outputs. The ideas of knowledge distillation were extended to segmentation tasks by \citet{liu2019structured} and \citet{xie2018improving}. They introduced additional losses which take structural consistency into consideration.

Realizing the concern of unavailability of large datasets during knowledge distillation, mainly due to privacy or security concerns, there have been recent works on the paradigm called data-free knowledge distillation. \citet{lopes2017data} were the first to introduce this paradigm. Along with the model, they store statistics of activation records of hidden units as metadata. These activations can be used to reconstruct samples from the dataset. A completely data-free approach was proposed by \citet{nayak2019zero}. To generate representative samples, they sample output vectors of the softmax layer from a mixture of Dirichlet distributions with carefully selected parameters to ensure diversity. Both these methods work on the principle of activation maximisation. \citet{addepalli2019degan} used an adversarial framework named Data-Enriching GAN to extract representative samples from a trained classifier. \citet{micaelli2019zero} also make use of an adversarial framework. They achieve this by training an adversarial generator to search for images on which the student poorly matches the teacher, and then using them to train the student. For dense prediction tasks like segmentation, activation maximisation approaches have limitations. For our proposed approach, we use the principle of an adversarial training framework.

\subsection{Data-Enriching GAN (DeGAN)}
The Data-Enriching GAN \cite{addepalli2019degan} proposes a novel training framework for performing Knowledge Distillation in the absence of data. It introduces the concept of a proxy dataset, which can be used as an alternative to the true dataset for performing Knowledge Distillation. The proxy dataset may lack diversity, may be imbalanced and lack the representations of the true dataset. To extract relevant representations from the true dataset, the training framework introduces a three player adversarial game between the Generator, Discriminator and the Classifier. While the discriminator tries to bring the distribution of the generated data closer to that of the proxy dataset, the classifier tries to bring in features specific to the true dataset. To our avail, we replace the classifier by a segmentation network.

\section{Approach: Good things come in threes}

To extract samples from a trained segmentation network, we employ a generative approach similar to \citet{addepalli2019degan}. For the generative samples to be effective for downstream tasks like knowledge distillation, they should closely match the distribution of the data on which the network was trained. For the samples to be representative of the data, we formulate that the samples should satisfy the following three conditions:

\begin{enumerate}
\item Samples should be indistinguishable from natural images.
\item Samples should be confidently classified by the network.
\item Samples should be diverse and represent all classes. 
    
\end{enumerate}

In the following subsections, we elaborate on these three how these three conditions are achieved. Our experiments show that these three conditions are sufficient to extract knowledge from a trained segmentation network which are useful for knowledge distillation.

\subsection{Proxy Images: Generate real-looking samples}
Approaches based on activation maximization have shown that data-free knowledge distillation is possible even if the samples generated do not look natural. These approaches work well in classification, as the task is to output a probability vector. In segmentation networks structural consistency of objects in segmentation maps is crucial. This makes it important for the samples to look natural. Since the true dataset is not available, we use proxy images. Proxy images are unlabelled natural images that are collected for the task which is to be performed. Such images can be collected from the internet, or by performing the inexpensive task of capturing new images. To ensure that the task of collection of proxy images is easy, we allow that the images need not contain all classes on which the segmentation network was trained. For eg. it is fine if none of the proxy images contain a bicycle even if the true dataset had labelled images for it.


Several generative approaches have been explored to generate samples \cite{oord2016pixel, kingma2013auto} from an image distribution. We make use of a Generative Adversarial Network (GAN) \cite{goodfellow2014generative} framework to generate real-looking images. GANs are known to be difficult to optimise for large images. To this end, we make use of Deep Convolutional GAN (DCGAN) \cite{radford2015unsupervised} which has shown to be stable for large images.

The training framework of DCGAN consists of a fully convolutional Generator $G(\bigcdot)$, and a fully convolutional Discriminator $D(\bigcdot)$. The Generator samples a vector from the latent space, defined by the probability distribution $p_z(z)$, and generates an image sample $G(z)$. The Discriminator takes either the generated images as input, or a real image $x$ sampled from the distribution $p_{data}(x)$ of the proxy dataset. The adversarial losses are given as:

\begin{equation}
    L_D^{SGAN} = -E_{x\sim p_{data}(x)}[log(sigmoid(D(x)))] -E_{z\sim p_z(z)}[log(1-sigmoid(D(G(z))))]
\end{equation}
\begin{equation}
        L_G^{SGAN} = -E_{z\sim p_z(z)}[log(1-sigmoid(D(G(z))))]
\end{equation}




The Generator is optimized to generate images that match the distribution of proxy images. The generated samples may not lie in the image manifold of the true dataset. By monitoring the softmax output of these images passed through a trained segmentation network, two observations can be made. The predictions made are not confident, and some prediction classes are underrepresented. The DeGAN framework allows us to ensure these constraints in the form of entropy loss and diversity loss.

\subsection{Entropy Loss: Samples should be confidently classified}

For images that fall in the true data distribution, the segmentation network produces a confident prediction for each pixel in the image. The proxy images are not guaranteed to follow the true data distribution. If the softmax output is confident towards a particular class, the entropy of the softmax distribution will be low.  To add this as a constraint, the entropy of the softmax distribution of each pixel is minimised. We make use of a pretrained segmentation network $F(\bigcdot)$. The segmentation network includes a softmax layer as the last layer. It is trained on a task containing $K$ classes. The generated images $G(z)$ are passed through the network to obtain the segmentation maps $s$ as formulated in equation (3). For a segmentation map of size WxH, we formulate the entropy loss as

\begin{equation}
s = F(G(z)), \quad
L_{G, entropy} = E_{z\sim p(z)}[\frac{1}{W\bigcdot H} \sum_{i=0}^W \sum_{j=0}^H \{- \sum_{k=0}^{K} s_{i,j,k}log(s_{i,j,k})\}]
\end{equation}

\subsection{Diversity Loss: Samples should represent all classes}

\begin{figure}
\begin{subfigure}{.3\textwidth}
  \centering
  \includegraphics[width=.8\linewidth]{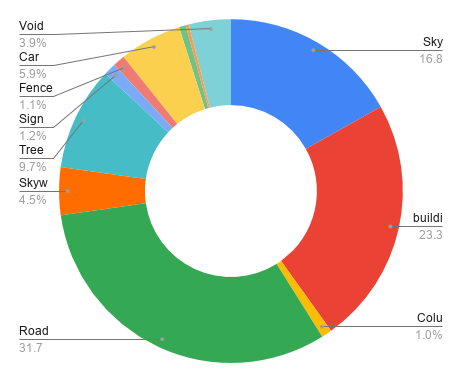}  
  \caption{CamVid dataset labels}
  \label{fig:pix-first}
\end{subfigure}
\begin{subfigure}{.3\textwidth}
  \centering
  \includegraphics[width=.8\linewidth]{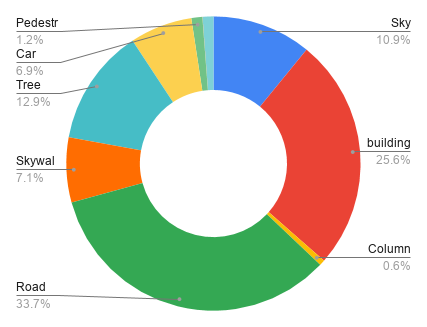}  
  \caption{DCGAN generated images}
  \label{fig:pix-second}
\end{subfigure}
\begin{subfigure}{.3\textwidth}
  \centering
  \includegraphics[width=.8\linewidth]{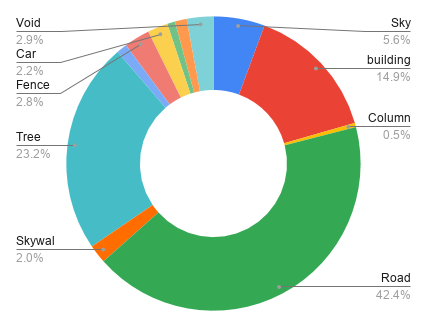}  
  \caption{DeGAN generated images}
  \label{fig:pix-third}
\end{subfigure}
\caption{The pie charts represent the distribution of pixels over classes. Fig. \ref{fig:pix-first} shows the distribution among the ground truth labels. This is comapared to the distribution of pixels in generated methods in Fig. \ref{fig:pix-second} and Fig. \ref{fig:pix-third}. DeGAN generates a much diverse distribution compared to DCGAN.}
\label{fig:pixels}
\end{figure}

There is no way to ensure that the proxy images cover all the classes. \citet{Kdoriginal} showed that knowledge about these classes can be extracted from a pretrained network. For achieving this, we pass a batch of generated images through a segmentation network. To ensure that all sets of classes are present across the segmentation maps, the entropy of the average prediction of pixels of images in a batch can be maximised. The expected softmax output of pixels over a batch of images is given by $w$. The diversity loss can be formulated as:

\begin{equation}
    w_k = E_{z\sim p(z)}[\frac{1}{W\bigcdot H} \sum_{i=0}^W \sum_{j=0}^H s_{i,j,k}], \quad
    L_{G,diversity} = \sum_{k=0}^{K}w_klog(w_k)
\end{equation}

\paragraph{Weighted Diversity Loss}The loss given above obtains a minimum when all classes are equally distributed across a batch. Figure \ref{fig:pix-first} shows the distribution of labels across the CamVid dataset. Certain classes are underrepresented by the dataset, eg. fence. The loss $L_{G, diversity}$ makes it difficult for the network to produce underrepresented classes as output. To ensure that these classes get more weight a new formulation for diversity loss is given as follows - 

\begin{equation}
    L_{G,weighted} = \sum_{k=0}^K \{\frac{1}{w_k}(\frac{1}{K} - w_k)^2 \}
\end{equation}

This loss can be interpreted as minimising the L2 distance between the distribution of pixels across a batch and the uniform distribution weighted by its under-representation. The effect of this loss function is shown in figure \ref{fig:pixels}.

\subsection{Architecture: Putting it all together}

\begin{figure}
  \centering
  \includegraphics[scale=0.4]{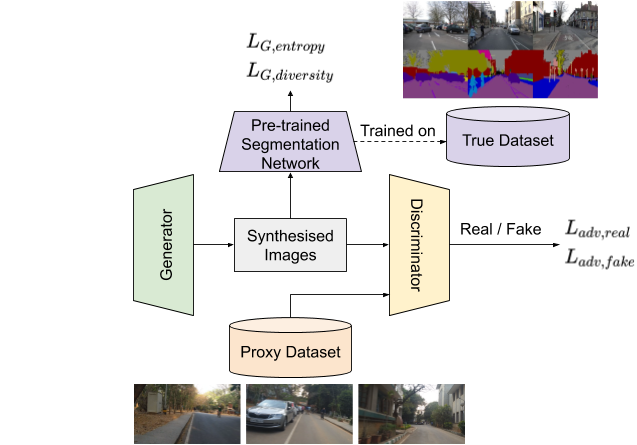}
  \caption{Data-Enriching GAN architecture}
  \label{fig:degan}
\end{figure}

As introduced by \citet{addepalli2019degan}, the training framework is a three player game between a Generator, a Discriminator and a Segmentation network. This is shown in figure \ref{fig:degan}. The Generator is jointly trained to minimize adversarial loss, diversity loss and entropy loss. The total loss is given as follows. $\lambda_d$ and $\lambda_e$ are the weights corresponding to diversity and entropy loss respectively.

\begin{equation}
    L_G^{total} = L_G^{GAN} + \lambda_d L_{G, diversity} + \lambda_e L_{G, entropy}
\end{equation}

\subsection{Mixed-data Knowledge Distillation}

Segmentation networks perform dense predictions. For achieving better performance, they view the images at multiple scales and thus achieve better object localization. In this sense, segmentation networks rely on object details to make correct predictions. Generative methods like GANs have difficulty in generating object details. This reduces the effectiveness of these images as representative samples. On the other hand, if proxy images are used as it is, they lack the diversity to learn all classes correctly. To achieve both quality as well as diversity, we construct mini-batches of size $n_{batch}$ containing images from the proxy dataset ($x_{dataset}$) and images generated by the generator ($x_{gen}$). To test out various mixing ratios, we define hyperparameters $\alpha$ and $\beta$. Mathematically, 

\begin{gather}
x_{\alpha} \subset X_{dataset}, \quad x_{\beta} \subset X_{gen} \\
s.t. |x_{\alpha}|=\alpha; |x_{\beta}|=\beta;
\alpha+\beta=n_{batch} \\
x_{mix} = \{ x_{\alpha}, x_{\beta} \}
\end{gather}

\paragraph{Knowledge Distillation Objective} We define the knowledge distillation objective in a data-free scenario as minimizing the KL-Divergence between the predictions of the teacher $T(\bigcdot)$ and the student $S(\bigcdot)$.

\begin{equation}
\begin{split}
    L_{kd} & =  D_{KL}(T(x) || S(x)); \\ 
    & \text{where} 
D_{KL}(T(x)||S(x)) = \sum_i t^{(i)}log(t^{(i)}/s^{(i)})
\end{split}
\end{equation}



\section{Experiments}

We evaluate our proposed approach of data-free knowledge distillation for semantic segmentation on the CamVid \cite{brostow2009semantic} dataset, which is a benchmark for autonomous driving. Though several other popular benchmarks are available, we prefer this dataset for its small size.

\subsection{Dataset Setup}

\paragraph{True Dataset}
The CamVid \cite{brostow2009semantic} dataset contains images of resolution 360x480.  It has 12 classes consisting of 367 training images and  101 validation images. To train a GAN with reasonable computational cost, we resize the images to 128x170. Since a GAN generates square images, we create training samples by taking center crops of size 128x128 from the resized images and pass it to the segmentation network. The same transformations are applied to validation sets as well. This allows for a fair comparison of our approach with baseline approaches.

\paragraph{Proxy Dataset}
Cityscapes \cite{cordts2016cityscapes} contains 2975 training images. We discard the labels and use these images as a proxy dataset for training the generator. We argue that though it is difficult to obtain labelled training data, collection of images is an easier task. So having a large amount of images in our experiments is acceptable.

\subsection{Segmentation}

We use DeeplabV3 \cite{chen2017deeplab} architecture for segmentation. For the teacher network we use a ResNet-50 backbone and an output stride of 8. The teacher network has $\sim$$39$ million parameters. For the student network we use a MobileNetV2 backbone. It has $\sim$$5$ million parameters. Having an output stride of 8 adds huge memory overhead, which is undesirable for mobile devices, so we use an output stride of 16 for the student network. The teacher network achieves a mean IoU of 56.08\%  while the student attains 49.16\% as shown in Table \ref{kd}.

\paragraph{Learning Policy}
All segmentation networks were trained for 120 epochs, with dropout (p=0.5), weight decay of 0.0005 and an initial learning rate of 0.001. The learning rate was multiplied by a factor of 0.1 after every 40 epochs.

\subsection{Data-Enriching GAN}

The hyper parameters for training the DCGAN  architecture is similar to \citet{radford2015unsupervised}. The learning rate was set to 0.0002, and Adam optimizer was used with $\beta_1$, $\beta_2$ as 0.5 and 0.999 respectively. All GANs were trained for $\sim$ 150K iterations (ie. 50 epochs for Cityscapes having $\sim$ 3000 images). Initially a batch size of 32 was used. Faster convergence was observed when a batch size of 16 was used. The GANs show improvement in quality of images even after training for 500 epochs, especially for larger image sizes. It was avoided for ease of hyperparameter tuning.

\begin{figure}
  \centering
  \includegraphics[scale=0.4]{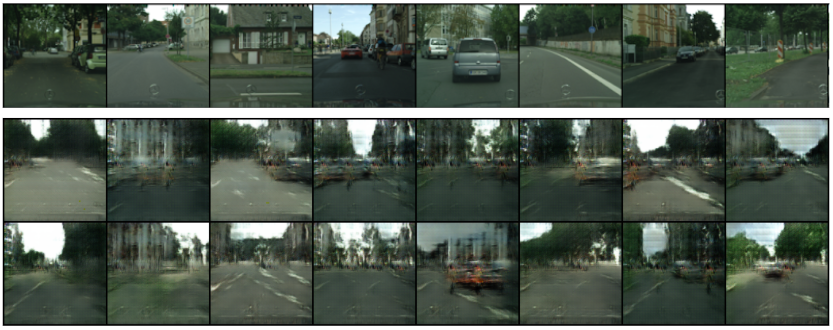}
  \caption{Top Row: Some example images for the Cityscapes dataset. Bottom two rows: Samples generated by DeGAN}
  \label{fig:samples}
\end{figure}

\begin{figure}
  \centering
  \includegraphics[scale=0.4]{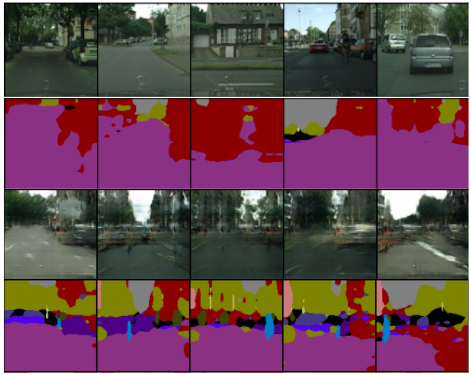}
  \caption{Top to Bottom: 1:Images from Cityscapes dataset. 2:Segmentation maps obtained by passing through a segmentation network trained on CamVid dataset. 3: Samples generated by DeGAN. 4: Segmentation maps obtained by passing the generated samples through the segmentation network which look more semantically meaningful. }
  \label{fig:qual}
\end{figure}

Some samples of generated images from GANs are shown in figure \ref{fig:samples}. For tuning of hyperparameters $\lambda_e$ and $\lambda_d$, we chose values similar to those used by \citet{addepalli2019degan}. Some qualitative results of the effect of hyperparameters on segmentation maps is shown in figure \ref{fig:qual}.

\subsection{Knowledge Distillation}

We perform various experiments to show upper bounds and effectiveness of our technique in Table \ref{kd}. At first, we show the performance of the teacher and student with labelled data. The third row shows the performance of distillation without using labels. Then we train a generator using DCGAN and DeGAN architectures We show a performance   improvement of 6.39\% using our formulation of weighted diversity loss and mix-data knowledge distillation.

\begin{table}
  \caption{Knowledge Distillation for Segmentation.}
  \label{kd}
  \centering

 \begin{tabular}{ccc}
    \toprule
    Model   & Mean IoU & Pixelwise Accuracy \\
    \midrule
    Teacher (Resnet-50 backbone)    & 56.08  & 90.02 \\
    Student (Mobilenetv2 backbone)     & 49.16 & 88.51 \\
    True Dataset  & 49.97 & 88.74 \\
    \cmidrule(r){1-3}
    Proxy dataset & 29.43 & 70.52 \\
    DCGAN   & 22.74 & 66.91   \\
    DeGAN   & 23.59 & 67.70   \\
    DeGAN + weighted diversity & 25.95 & 73.11 \\
    DeGAN + mixed data & 28.48 & 76.23 \\
    DeGAN + weighted diversity + mixed data & \textbf{30.52} & 75.02 \\
    \bottomrule
  \end{tabular}
\end{table}

\begin{table}
  \caption{Ablation Study of Diversity and Entropy losses}
  \label{entdiv}
  \centering
  \begin{tabular}{cc|cc|cc}
    \toprule
    \multicolumn{2}{c|}{Hyperparameters} & \multicolumn{2}{c|}{Diversity Loss} & \multicolumn{2}{c}{\textbf{(ours) Weighted Diversity}} \\
    $\lambda_e$ & $\lambda_d$ & Mean IoU & Pixelwise Accuracy & Mean IoU & Pixelwise Accuracy \\
    \midrule
    0 & 0 & 22.74 & 66.91 & 22.74 & 66.91 \\
    0 & 10 & 21.72 & 64.88 & \textbf{25.95} & 73.11 \\
    10 & 0 & 23.59 & 67.70 & 23.59 & 67.70 \\
    10 & 10 & 20.41 & 64.27 & 18.11 & 59.84 \\
    5 & 10 & 23.04 & 66.28 & 25.54 & 69.37 \\
    \bottomrule
  \end{tabular}
\end{table}


\paragraph{Ablation Studies}
Table \ref{entdiv} shows the efffect of hyperparameters $\lambda_e$ and $\lambda_d$ on the performance of knowledge distillation. $\lambda_e=0$ and $\lambda_d=10$ appears to be a good set of hyperparameters for our weighted diversity loss. We attain a performance improvement of $2.36\%$ using our approach.

\begin{figure}
  \centering
  \includegraphics[scale=0.4]{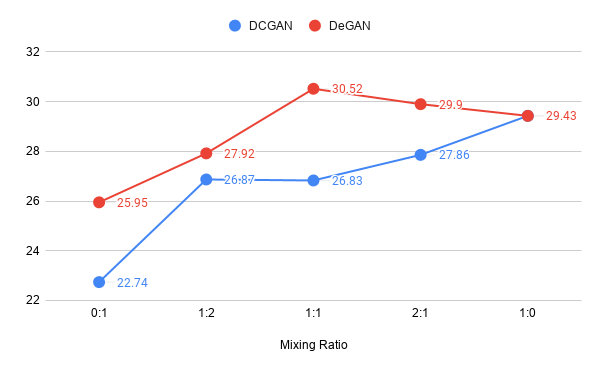}
  \caption{Ablation study for mix-data knowledge distillation for various mixing ratios. }
  \label{fig:mixresults}
\end{figure}

\paragraph{Mix-data Knowledge Distillation}
Figure \ref{fig:mixresults} shows that using only generated images (i.e. mixing ratio of 0:1) performs poorly compared to if we simply use the proxy images (i.e. mixing ratio of 1:0). We show results for various mixing ratios ($\alpha:\beta$) Using DCGAN generated images and mixing them does not help to improve performance. We show that in a DeGAN framework, the mix-data approach improves performance over just using the generated images or just using proxy images.

\section{Conclusion}

We demonstrated the extensibility of the DeGAN framework to segmentation tasks. We scaled the DeGAN networks for large size images. To tackle the underrepresentation of classes in segmentation networks, we proposed a new weighted diversity loss. For improving the performance knowledge distillation by ensuring quality and diversity we made use of mix-data batches. We presented a new pipeline for model compression which does not raise privacy concerns and can be used by large corporations to release their models.

\bibliographystyle{plainnat}
\bibliography{main}






\end{document}